\documentclass[conference,compsoc]{IEEEtran}[2015/08/26]

\usepackage[utf8]{inputenc}
\usepackage{graphicx}
\usepackage[ngerman,english]{babel}
\usepackage{csquotes}
\usepackage{url}
\usepackage{booktabs}
\usepackage[square,comma,numbers,sort]{natbib}
\usepackage{hyperref}
\usepackage[capitalise,nameinlink]{cleveref}

\addto\extrasenglish{\languageshorthands{ngerman}\useshorthands{"}}
\makeatletter
\g@addto@macro{\UrlBreaks}{\UrlOrds}
\makeatother
\hypersetup{hidelinks,colorlinks=true,allcolors=black,pdfstartview=Fit,breaklinks=true}
\crefname{lstlisting}{\lstlistingname}{\lstlistingname}
\Crefname{lstlisting}{Listing}{Listings}

\begin{document}

\title{Bio-Measurements Estimation and Support in\\Knee Recovery through Machine Learning}

\author{
    \IEEEauthorblockN{João Bernardino}
    \IEEEauthorblockA{
        DEI - Faculty of Engineering\\
        University of Porto\\
        Porto, Portugal
    }
    \and
    \IEEEauthorblockN{Luís Filipe Teixeira}
    \IEEEauthorblockA{
        INESC TEC and\\
        DEI - Faculty of Engineering\\
        University of Porto\\
        Porto, Portugal
    }
    \and
    \IEEEauthorblockN{Hugo Sereno Ferreira}
    \IEEEauthorblockA{
        INESC TEC and\\
        DEI - Faculty of Engineering\\
        University of Porto\\
        Porto, Portugal
    }
}

\maketitle

\begin{abstract}
    Knee injuries are frequent, varied and often require the patient to undergo intensive rehabilitation for several months. Treatment protocols usually contemplate some recurrent measurements in order to assess progress, such as goniometry. The need for specific equipment or the complexity and duration of these tasks cause them to often be neglected. A novel deep learning based solution is presented, supported by the generation of a synthetic image dataset. A 3D human-body model was used for this purpose, simulating a recovering patient. For each image, the coordinates of three key points were registered: the centers of the thigh, the knee and the lower leg. These values are sufficient to estimate the flexion angle. Convolutional neural networks were then trained for predicting these six coordinates. Transfer learning was used with the VGG16 and InceptionV3 models pre-trained on the ImageNet dataset, being an additional custom model trained from scratch. All models were tested with different combinations of data augmentation techniques applied on the training sets. InceptionV3 achieved the best overall results, producing considerably good predictions even on real unedited pictures.
\end{abstract}

\section{Introduction} \label{sec:intro}
    A study published in 2006 revealed that, from 19.530 sports injuries documented over a 10-year period, close to 40\% were related to the knee \cite{Majewski:2006df}, making it one of the most commonly injured structures of the human body, especially in athletes. Unfortunately, in a great number of cases, full recovery takes several months, during which a physical therapy program is usually recommended.
    
    In order to assess the patient's evolution and efficiently conduct the rehabilitation, health professionals (physiatrists and physical therapists) must perform recurrent measurements. Goniometry is one of the most common, which consists of determining the knee's range of motion or flexion angle. In order to perform this accurately, some tool must be used. However, most of these tools are either not available at every physical therapy clinic or its usage is complex and time-consuming. This often causes them to be replaced by simple visual estimation which, while not precise, is extremely quick and pratical to execute. Thus,in most cases, the evaluation tends to be mainly qualitative, and consequently imprecise. This research is supported by the idea that an easier and quicker way to perform meaningful measurements would contribute to an increased amount of evaluations performed, as well as a deeper insight on the patient’s progress. 

    A novel solution is proposed, consisting of the creation of a mobile application capable of performing accurate measurement of the knee's range of motion. It should be a low-cost solution that is simple and quick enough to operate, so that physical therapists, even more that physiatrists, can use it frequently, as they are the professionals that most closely follow and deal with the patient during the treatment. Specifically, the app should be capable of estimating the joint's flexion angle using a common smartphone’s camera. The estimations should be executed in real-time, without requiring photography or any interaction by the professional other than pointing the camera, and the obtained results should promptly be presented on the device’s screen.

\section{State of the Art} \label{sec:sota}
    This section is dedicated to reviewing state of the art solutions for performing accurate goniometry on the knee joint, in the context of injury rehabilitation.

    \subsection{Traditional Solutions}
        While, as previously mentioned, this kind of measurement is often based on simple visual estimation, several tools and techniques have been developed through the years and are currently used by many health professionals.

        \subsubsection{Radiography}
            The technique generally accepted as reference consists of direct observation of the bones in a radiograph \cite{Edwards:2004fg}. However, due to radiation exposure, this method cannot be used frequently.

        \subsubsection{Universal goniometer}
\begin{figure}[ht]
                \centering
                \includegraphics[width=0.35\textwidth]{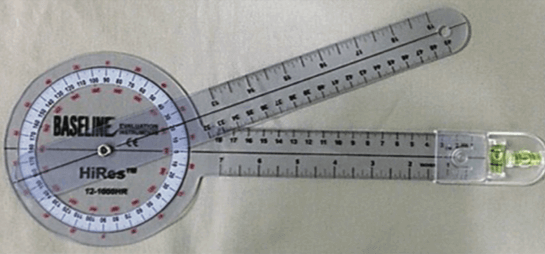}
                \caption[Universal goniometer]{Universal goniometer \cite{Roach:2013wr}}
                \label{fig:universalgoniometer}
            \end{figure}

            This tool, depicted in \cref{fig:universalgoniometer}, is the one most commonly used. Research indicates that, with more than 30 degrees of knee flexion, there is no significant difference between the results obtained using this method and the radiographic one \cite{Enwemeka:1986vy}. While it is considered one of the most reliable solutions, it still has some disadvantages. Besides requiring a specific physical device, it must be manually manipulated by the health professional. This introduces some error, so it is advisable that all measurements on a patient are performed by the same person \cite{Hellebrandt:1949je}. Another disadvantage is the fact that this method is unusable in motion — the patient's leg must be stationary.

        \subsubsection{Digital protractor}
            This device works in a similar fashion as a bubble inclinometer, reporting the inclination of a surface. While the flexion angle cannot be directly measured with this tool, it can be determined through the inclination of the thigh and the lower leg.

    \subsection{Smartphone-based Solutions}
        Nowadays, smartphones play a key role in the daily routines of a large number of individuals. Most of these are packed with a diverse set of sensors, including digital cameras, magnetometers and accelerometers. This makes such devices quite suitable for a great variety of applications and purposes, and health care is no exception. Whether it aids the retrieval of information, supports diagnosis or provides any other kind of solution, the use of smartphones in medicine is growing at a quick pace \cite{Mosa:2012fk}. In fact,
        some smartphone-based solutions already exist for performing goniometry.

        \subsubsection{Using inertial sensors}
\begin{figure}[ht]
                \centering
                \includegraphics[width=0.48\textwidth]{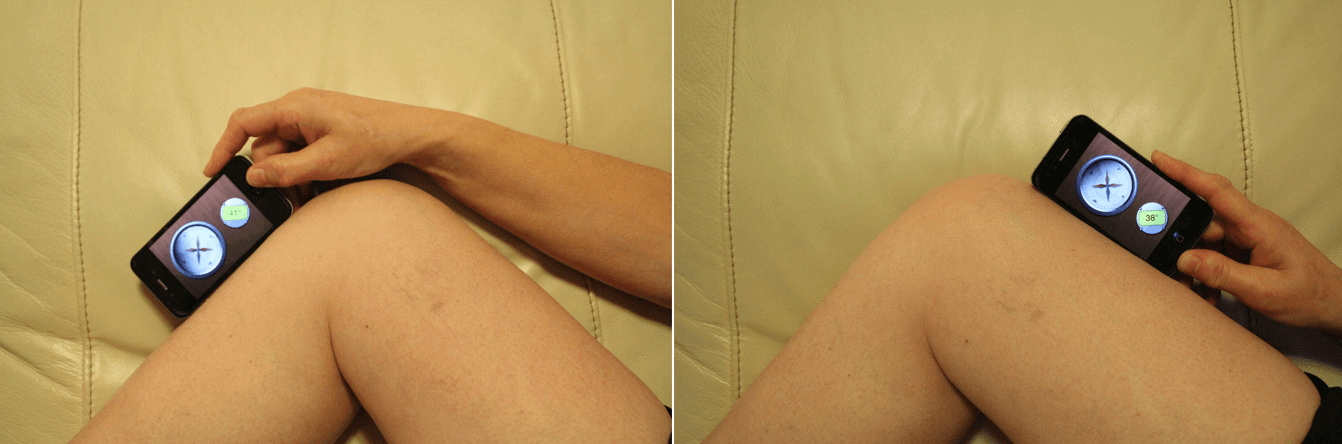}
                \caption{Knee goniometry using a gyroscope-based application \cite{Jenny:2013bw}}
                \label{fig:angleapp}
            \end{figure}

            Most smartphones contain an inertial sensor called gyroscope. This component relies on Earth’s gravity to determine the orientation, and it is usually calibrated to read a value of 0º when the device is on a horizontal position. Thus, a gyroscope-equipped smartphone can provide the same functionality as a digital inclinometer, making it suitable to perform goniometry. Again, this technique also requires the measurement of two inclinations (thigh and lower leg) before determining the joint’s angle, as depicted in \cref{fig:angleapp}.

            A study evaluating a similar mobile application, performing knee goniometry at six different leg positions (full extension, 30º, 60º, 90º, 110º and maximal flexion), indicates a slightly higher accuracy of this measurement technique when compared to conventional ones \cite{Jenny:2013bw}. Previous research demonstrates that this approach is precise enough to be frequently used in a clinical environment. Moreover, there are multiple similar applications available for free, making this an inexpensive solution for performing knee goniometry (excluding the cost of the smartphone).

        \subsubsection{Using photography}
            DrGoniometer\cite{Vercelli:2017dk} is a mobile application that is capable of measuring the angles of the patient’s articulations. It uses a photography-based approach, thus being appropriate for static joint angle measurement only. As it was earlier mentioned, in order to get an accurate estimate the camera’s point of view should be parallel to the plane defined by the leg \cite{Ferriero:2013iy}. Doing so will avoid errors induced by perspective distortion. It’s worth noting that this rule is not specific to this application, but rather any photography-based angle measurement system.

            \begin{figure}[ht]
                \centering
                \includegraphics[width=0.25\textwidth]{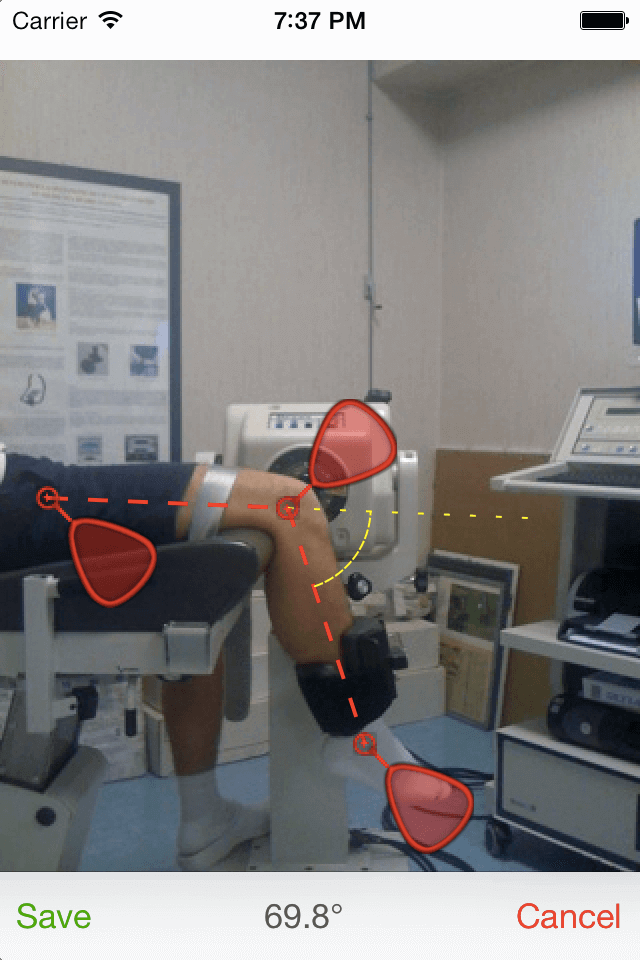}
                \caption{Knee ROM measured with DrGoniometer}
                \label{fig:drgoniometer}
            \end{figure}

            In this particular case, after taking a picture, three markers must be dragged to the appropriate positions. Recommended reference points are the center of the knee, the hip and the ankle \cite{Ferriero:2013iy}. Having these three markers positioned, the application presents two dashed lines representing the angle of the joint, and the measured value is visible on the bottom of the screen. Using the coordinates of the markers, the angle calculation is based on simple trigonometry. Studies have been published demonstrating this solution's reliability and validity, not only for measuring knee range of motion, but also other joints of the human body \cite{Otter:2015gm}. The results reveal measurement errors similar to those of traditional measurement tools, such as the Universal Goniometer, making it a perfectly valid solution to be used by professionals \cite{Ferriero:2013iy}.

\section{Background} \label{sec:background}
    \subsection{Image Recognition}
        Computerized image recognition is essential in the context of this dissertation, as the proposed solution consists in using the smartphone’s camera to accurately perform body measurements. The following sections review theoretical concepts on this topic that are related to the solution being proposed.
        
        \subsubsection{Computer vision}
            This is the field of computer science that focuses on the task of enabling computers to process and understand digital images and videos, usually with the ultimate goal of automating human tasks. It involves automatic extraction and analysis of features from the digital data, such as patterns, and a consequent retrieval of relevant information from them, which is essential for achieving automatic image recognition. Many modern computer vision techniques rely on machine learning algorithms.

        \subsubsection{Machine learning}
            This is a branch of artificial intelligence which focuses on using techniques that allow the computers to automatically update their behavior in order to improve its performance on a given task. As the name suggests, machine learning algorithms are based on the definition of learning, which can be described as the ability to generalize from prior experience. By analyzing a given set of labeled data, commonly referred to as the \textit{learning dataset} or \textit{training dataset}, the machine should be able to perform accordingly on new examples it has never seen before. Thus, the learning dataset should be representative and of considerable dimension, which is not always easy to obtain. This is one of the biggest challenges of every machine learning approach: its need for data \cite{Langley:1988jk}. Machine learning algorithms can be grouped in several categories, including deep learning, which has recently become one of the most popular and successful approaches regarding computer vision.

        \subsubsection{Deep learning}
            Despite being a relatively young field, remarkable results have been achieved by resorting to this type of technique. It differs from other machine learning algorithms mainly for its use of hierarchical structures, representing data with multiple levels of abstraction \cite{Sinha:2018vm}. Among all deep learning strategies, the use of convolutional neural networks (CNN) is one of best in terms of image recognition.

        \subsubsection{Convolutional neural networks}
            These consist of layered structures of connected neurons, where each layer receives input data and passes a transformed output to the next layer, until the end is reached. A common CNN has three types of neural layers: convolutional layers, pooling layers and fully-connected layers \cite{Voulodimos:2018bf}. In convolutional layers, multiple kernels are used to extract feature maps from parts of the image. The purpose of pooling layers is the reduction of the spatial dimensions. The output volume is smaller than the input volume. This is achieved by combining multiple adjacent values into a single one. The strategy used for this process is what differs between different types of pooling layers. Convolutional and pooling layers account for most of a CNN's structure and are responsible for extracting and learning relevant features from the images. At the end of the network, there are usually a few fully connected layers. In contrast to the other two kinds of layers, here every neuron is connected to all the neurons of the next layer. These are responsible for the CNN's reasoning and producing the final output, or prediction. The way each layer transforms its input to its output is dependent on a number of parameters, commonly called \textit{weights}. As with any machine learning algorithm, CNNs need to be trained with labeled data, and in particularly high amounts. The training of a CNN consists of an interative process where, after running each image through all the layers, the weights are adjusted in an attempt to match the prediction with the expected outcome (the labels). To evaluate the quality of the predictions, CNNs use a loss function to compare them with the labels. On an untrained model, the initial weights are usually random, so it is expected that the results of the first iterations are not good. However, it is expected that the loss values progressively decrease, describing an exponential decay curve, as the weights become more properly tuned. Eventually, they should converge to a (preferably low) value, meaning that it's no longer improving and the learning process should stop \cite{Cao:2018gb}.

    \subsection{Transfer Learning}
        The requirement for large amounts of training data is one of the main disadvantages of convolutional neural networks, and the reason why they are not a viable solution in many cases. There is, however, a strategy to help circumvent this issue, called transfer learning. This consists of starting with a model pre-trained on a different dataset, instead of training from scratch \cite{Pan:2010dm}. Some popular models, created and trained during machine learning competitions, are publicly available. Most were trained on the ImageNet dataset. Transfer learning not only makes the training process faster and less resource-hungry, it also helps achieving good results with smaller datasets. While training, some layers of the CNN can be locked, which stops their weights from being adjusted. Alternatively, they can be left unlocked and, thus, the weights will be adjusted. This is commonly known as fine tuning \cite{Shin:2016cx}.

    \subsection{Synthetic Data in Deep Learning}
        Another workaround for the lack of large amounts of data, with which positive results have previously been demonstrated, is synthetic data generation. If using artificially generated images, dataset size should no longer be a problem. This, of course, is not the ideal approach and bring its own set of problems. Specifically, it might be tough to get the CNNs to generalize from what is learned on synthetic data to real data, which is the ultimate goal. The data generation algorithm must produce realistic samples that closely resemble reality in order to make this possible. Ekbatani et. al demonstrated a successful application of this strategy in a problem of counting the number of pedestrians in street photos \cite{KeivanEkbatani:gn}.
        
                \section{Problem Statement} \label{sec:problem}

                \subsection{Requirements}
                    The ultimate goal of this project is the development of a low-cost smartphone-based solution that aids physiatrists and physical therapists during the usually long rehabilitation process that follows a knee injury, allowing them to estimate knee range of motion with nothing more than a smartphone. It should consist of a simple application which, after pointing the device's camera at the patient's leg, immediately presents the knee flexion angle estimation on the screen. Furthermore, unlike existing solutions that rely on photography, this should not work only on previously captured images, but instead estimations shall be executed in real-time, allowing for dynamic measurement.
            
                \subsection{Solution Proposal}
                    Considering the requirements explained above, the biggest challenge is performing accurate image recognition. The system must be able to detect when the camera is pointing at a human leg, recognizing the shape and retrieving the desired information, which are the coordinates of three key points in the image (the knee, the center of the thigh and the center of the lower leg) required for the calculation of the joint’s angle. This should be achieved using convolutional neural networks. However, large datasets are required to adequately train these structures. Since there is no suitable dataset available, we will synthesize one, taking advantage of 3D modeling software reviewed earlier.
            
                \subsection{Evaluation}
                    Since the neural networks will be trained using synthetic data, the main goal is two fold: (i) to achieve high accuracy in synthetic images, and (ii) to accurately transfer/generalize these capabilities for real images. A non-artificial validation set will be created after capturing images of the leg, simulating a treatment environment, and manually labeling them by marking the expected coordinates to be compared with the neural network's predictions. At a more advanced stage, once the proposed solution is implemented and a functional mobile application is developed and ready to be used, it should be tested by actual physical therapists and physiatrists from a few local clinics, with the collaboration of Fábio Rocha, a physical therapist working as an external advisor in this dissertation. During the course of multiple therapy sessions with patients recovering from knee injuries, range of motion shall be measured using both the proposed solution and the traditional methods for comparison.

\section{Synthetic Data Generation} \label{sec:generation}

    \subsection{3D Model Creation}
\begin{figure}[ht]
            \centering
            \includegraphics[width=0.35\textwidth]{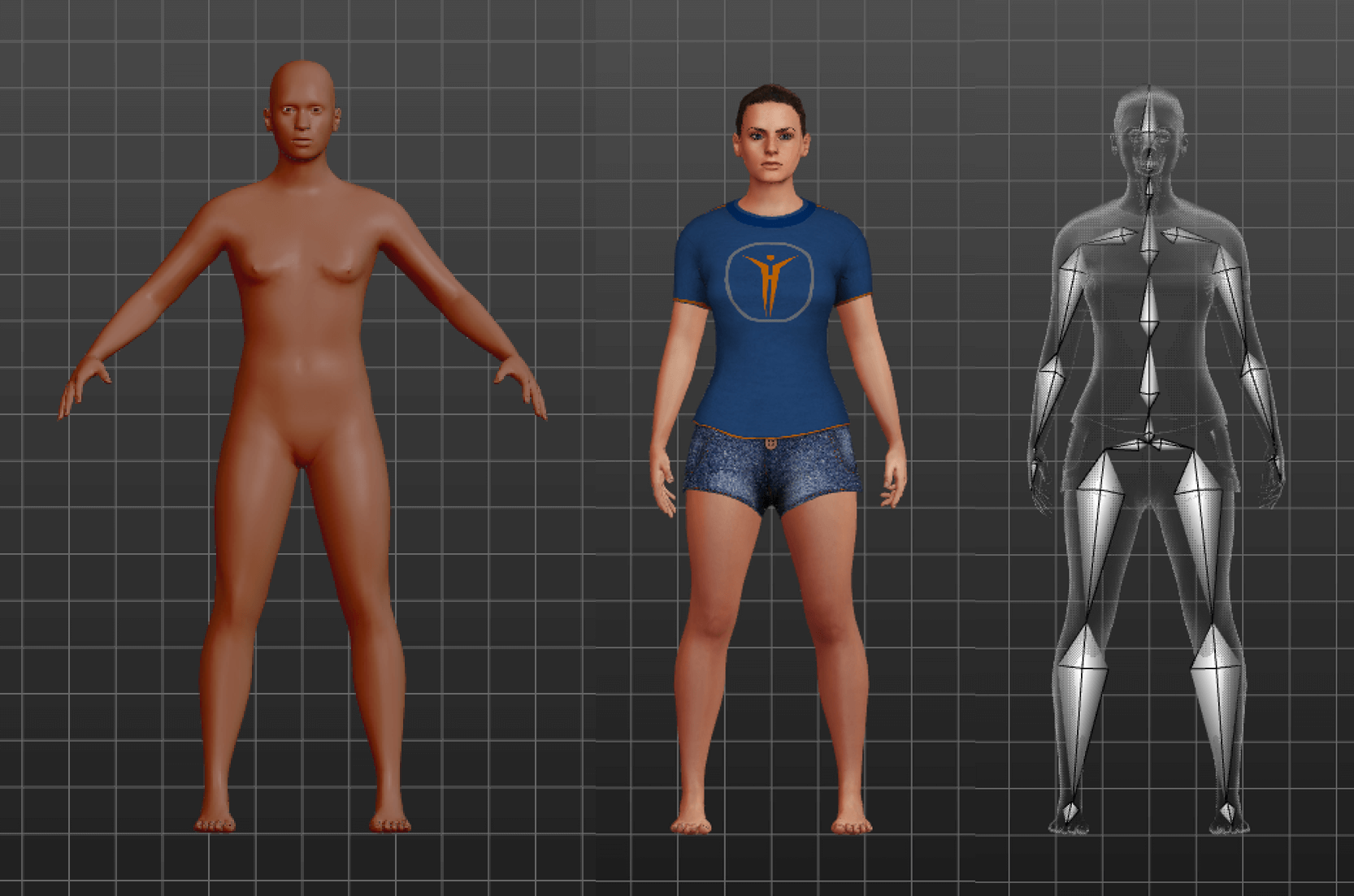}
            \caption{MakeHuman 3D models}
            \label{fig:makehuman}
        \end{figure}

        The first step of the data synthesis process was obtaining a 3D model of the human body. This was achieved using the open-source software MakeHuman\footnote{\url{http://www.makehumancommunity.org}}, a tool developed for the creation of three-dimensional characters. The model that was created not only has a realistic human body shape, but it is also rigged, meaning the 3D mesh is bound to a digital skeleton consisting of bones and joints, which makes it possible to pose the character according to the requirements. \Cref{fig:makehuman} shows the initial and final states of the developed model, as well as a diagram representing its skeleton.

    \subsection{Posing and Manipulation}
        The task of posing the model according to the requirements was executed in Blender\footnote{https://www.blender.org/}, as it allows the manipulation of individual bones of the character's skeleton. Thus, different angles of flexion were applied to the knee by simply rotating the lower leg bone. When a patient is lying on their back on a treatment table, flexing the knee also requires a flexion of the hip. Thus, a 1:2 relation between the hip's and the knee's flexion angles was defined, ensuring the heel was permanently aligned with the torso, as if it was resting on the treatment table. At this point, an image resolution of 200x150 pixels was established, taking into account characteristics of the CNN models used in the next step. To ensure that the entire was captured by the camera, regardless of the angle of flexion, the center of the image is an invisible point whose coordinates permanently match the centroid of the triangle described by the knee, the hip and the ankle. The scene also contains a light source positioned above the model to create appropriate lighting conditions. Additionally, in order to obtain more diverse data, multiple different-toned skin textures were used.

    \subsection{Automated Data Generation}
        With the 3D scene set up, the data generation task was automated by taking advantage of Blender's Python API. A simple script was developed to repeatedly render images from the scene's camera, while randomly applying different flexion angles between each iteration. For every rendered image, the program also writes to a comma-separated values (CSV) file the coordinates of the three points of the leg necessary for determining the flexion angle — the centers of the thigh, the knee and the lower leg. Since Blender provides coordinates in the range [0,1] relative to the top-left corner, these are converted to absolute values by multiplying each value by the corresponding image dimension. Additionally, they are transposed to a coordinate system where the origin is the top-left corner of the image, as commonly observed in computer graphics. In order to match each image with its set of coordinates, its filename, which is a numerical index (excluding extension), is also stored in the CSV file. Thus, the data is structured as shown in \cref{tab:csv-file}.

        \begin{table}[ht]
            \centering
            \caption{Example of data stored in the CSV file}
            \label{tab:csv-file}
            \begin{tabular}{|c|c|c|c|c|c|c|}
                \hline
                img & thigh\_x & thigh\_y & knee\_x & knee\_y & leg\_x & leg\_y \\ \hline
                0   & 62       & 74       & 98      & 71      & 137    & 73     \\ \hline
                1   & 64       & 69       & 98      & 61      & 132    & 73     \\ \hline
                2   & 85       & 62       & 99      & 30      & 114    & 64     \\ \hline
                3   & 70       & 67       & 98      & 45      & 129    & 67     \\ \hline
                ... & ...      & ...      & ...     & ...     & ...    & ...    \\ \hline
            \end{tabular}
        \end{table}

        Within the Python script, some parameters can be adjusted in order to configure the data generation process. These include the number of samples to be generated, the range of possible knee flexion angles and also a maximum rotation offset, relative to the patients leg, for the camera's position. This offset is used because, in a real-case scenario, it is virtually impossible for the therapist to set the camera in an exact perpendicular direction to the plane defined by the patient's leg. Thus, at each iteration, not only a random flexion angle is computed, but also a rotation for the camera. This rotational range should, however, be small, as significant rotations cause perspective distortion, making it impossible to accurately measure the real angle in the image.

        \begin{figure}[ht]
            \centering
            \includegraphics[width=0.45\textwidth]{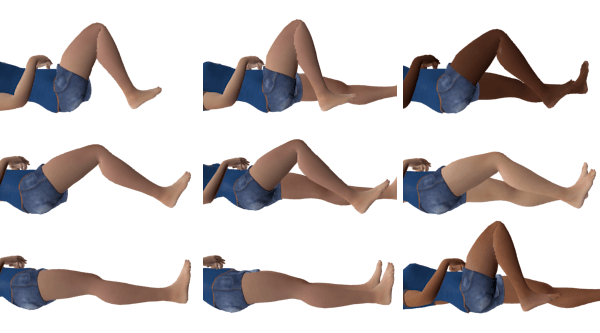}
            \caption{Examples of synthesized data}
            \label{fig:datasets}
        \end{figure}

        \Cref{fig:datasets} shows different types of samples that were obtained using this script: in the first column, only the front leg is visible (an attempt to simplify the deep learning process); in the middle one, both legs are visible. However, while the front leg is posed with random flexion angles, the other leg is straight in a resting position. And in the last column, also both legs are visible, but a different skin texture is randomly chosen for each sample. All images have a fully transparent background and were exported in Portable Network Graphic (PNG) format.

\section{Image Recognition} \label{sec:recognition}
    This part of the project was implemented using the Python 3.6 programming language, along with some useful frameworks and libraries. The core of this solution is built using Keras 2.2.0, a popular deep learning Python library, running on top of Google's TensorFlow 1.8.0. This stack allows the creation of custom CNN models, as well as the use of some pre-trained architectures, such as MobileNet, DenseNet, VGG16 and InceptionV3. The management and manipulation of images and data structures was supported by NumPy and Pillow, two Python libraries widely used in scientific computing. All training tasks were executed on a dedicated nVidia GeForce GTX 1080 graphics processing unit.

    \subsection{Data Augmentation}
\begin{figure}[ht]
            \centering
            \includegraphics[width=0.48\textwidth]{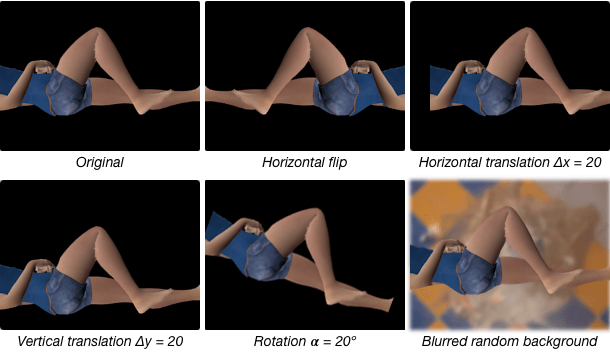}
            \caption{Examples of the transformations applied to the dataset}
            \label{fig:transformations}
        \end{figure}

        While it is easy to synthetically produce a large amount of data, introducing enough diversity to enable the neural networks to generalize what is learned from this data to real-life examples is a challenge. With that goal in mind, the following image augmentation techniques were applied to the dataset: horizontal flipping with probability of 50\%, horizontal translations with $ \Delta x \in [-20,20] $, vertical translations with $ \Delta y \in [-10,30] $, rotations with $ \alpha \in [-30^\circ,30^\circ] $ and, lastly, the replacement of the transparent background with one of 1491 random pictures from a diverse third-party dataset \cite{Jegou:2008il} to which a gaussian blur filter was previously applied. The geometric transformations should improve the CNNs performance on images where the leg is in different positions or orientations. The backgrounds are changed in an attempt to lead the CNNs to only recognize and consider patterns of the actual patient's leg as relevant. \Cref{fig:transformations} shows examples of the augmented dataset.

        If analyzed from left to right, the image points which the CNN is supposed to estimate are always in the same order (first the thigh, then the knee and finally the lower leg), with the single exception of horizontally flipped samples. This inconsistency hurts the training process, as it possibly leads to situations where the CNN predicts three correct pairs of coordinates, but not in the expected order. For instance, if the expected output was $[(1,0),(0,0),(-1,0)]$ and the network predicted $[(-1,0),(0,0),(1,0)]$, this would be evaluated as a bad prediction, even though it is not. Thus, whenever a horizontal flip is applied, the labels of the first and third points are swapped. This way we ensure the labels are always consistently ordered.

    \subsection{CNN Architectures}
\begin{figure}[ht]
            \centering
            \includegraphics[width=.48\textwidth]{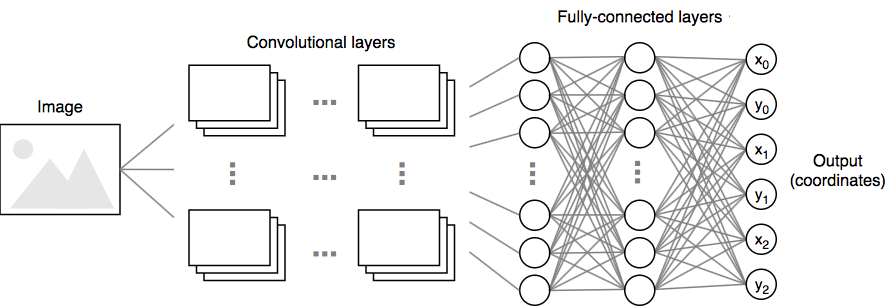}
            \caption{Diagram representing the required CNN structure}
            \label{fig:cnn-sketch}
        \end{figure}

        Three different CNN architectures were selected to test the proposed solution. The first two were InceptionV3\cite{Szegedy:cv} and VGG16\cite{Simonyan:2014ws}, both pre-trained on the ImageNet dataset, thus using transfer learning. Using these models required a slight modification. The first layer was updated to deal with the 200x150 resolution of the dataset, which was decided considering the minimum resolution required by most popular CNN architectures. And while we need the CNNs to output a set of continuous values (a regression problem), these two models were designed for classification problems. This required the final, fully-connected layer to be replaced with a new one with exactly 6 neurons and no activation function. The diagram in \cref{fig:cnn-sketch} loosely represents the structure of these three models (after adaptation).
        
        The third model, however, is a custom, much simpler CNN designed for this particular purpose, here referred to as \textbf{Eva}, which was trained from scratch with the synthetic dataset. It's full structure is represented in \cref{tab:eva}.

        \begin{table}[ht]
            \centering
            \caption{Architecture of the Eva CNN model}
            \label{tab:eva}
            \begin{tabular}{|l|l|l|}
                \hline
                \textbf{Layer Type} & \textbf{Output Shape} & \textbf{Parameters} \\ \hline
                Conv2D              & (None, 148, 198, 32)  & 896                 \\ \hline
                MaxPooling2D        & (None, 74, 99, 32)    & 0                   \\ \hline
                Conv2D              & (None, 71, 96, 64)    & 32832               \\ \hline
                MaxPooling2D        & (None, 35, 48, 64)    & 0                   \\ \hline
                Conv2D              & (None, 33, 46, 4)     & 2308                \\ \hline
                MaxPooling2D        & (None, 16, 23, 4)     & 0                   \\ \hline
                Flatten             & (None, 1472)          & 0                   \\ \hline
                Dropout             & (None, 1472)          & 0                   \\ \hline
                Dense               & (None, 100)           & 147300              \\ \hline
                Dense               & (None, 50)            & 5050                \\ \hline
                Dense               & (None, 6)             & 306                 \\ \hline
            \end{tabular}
        \end{table}

        To train a CNN using Keras, an optimization algorithm must be specified, and in this case it was RMSProp \cite{Wichrowska:2017vo}. Additionally, we also need to specify a loss function, which is used to assess the performance of the CNN and point the training process in the right direction. We implemented a custom function which calculates the \textbf{total euclidean distance} between the predicted $(x'_i,y'_i)$ and the expected $(x_i,y_i)$ three points, according to \cref{eq:loss}.

        \begin{equation} \label{eq:loss}
            \ell = \sum_{i=0}^{2} \sqrt{(x'_i - x_i)^2 + (y'_i - y_i)^2}
        \end{equation}

        Since a transfer learning strategy was applied, not all the layers were trained on the InceptionV3 and VGG16 models. The first layers of a CNN are responsible for recognizing basic patterns, such as edges. This \enquote{knowledge}, that was developed on a different dataset, is still relevant in this context. Thus, only the final portion of the CNN will be trained, which is where the actual prediction is generated. \Cref{tab:cnn-comparison} provides a comparison between the three models, showing the differences in dimension and proportion of trainable/non-trainable parameters.

        \begin{table}[ht]
            \centering
            \caption{Parameter distribution in the used CNNs. \# T and \# NT are the number of trainable and non-trainable parameters.}
            \label{tab:cnn-comparison}
            \begin{tabular}{|l|l|l|l|l|}
                \hline
                \textbf{Model} & \textbf{\# Total} & \textbf{\# T} & \textbf{\# NT} & \textbf{T Ratio} \\ \hline
                InceptionV3    & 23,907,110        & 10,605,510    & 13,301,600     & 45\%             \\ \hline
                VGG16          & 81,856,326        & 67,141,638    & 14,714,688     & 82\%             \\ \hline
                Eva            & 188,692           & 188,692       & 0              & 100\%            \\ \hline
            \end{tabular}
        \end{table}

    \subsection{Training Process}
\begin{figure}[ht]
            \centering
            \includegraphics[width=0.48\textwidth]{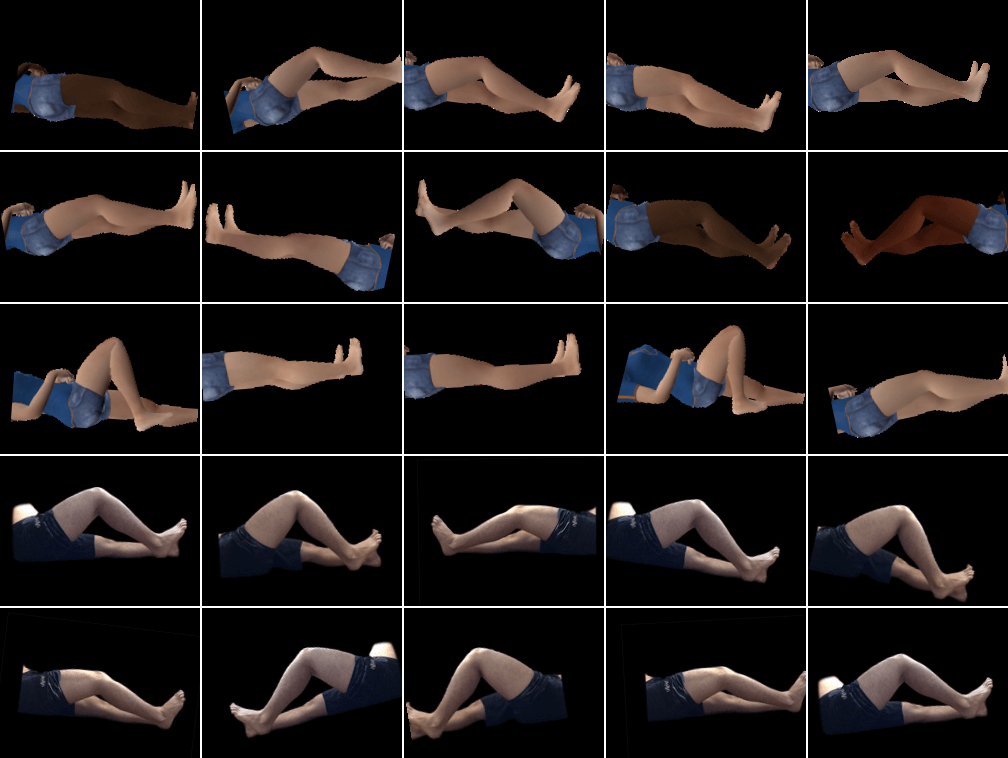}
            \caption{Samples from the validation sets}
            \label{fig:validation}
        \end{figure}

        In this phase of the project, the three selected CNN models are tested with different combinations of image augmentation techniques applied to the dataset. The training set has 3000 samples and is the same in all tests (only differs by the transformations applied on the fly). The validation set, however, is not exactly the same in each experiment. It is always composed of a total of 500 samples: 25 synthetic images with the original skin texture and only one leg visible; 425 synthetic images with varied skin textures and both legs visible; and 50 real pictures (originally only 3, to which similar image augmentation was applied in advance). The training sets differ in which image augmentation techniques were applied, matching the ones used for the training set of the respective experiment. \Cref{fig:validation} shows some samples used for validation, including real pictures (with the legs isolated over a fully transparent background).

\section{Results} \label{sec:results}
    All three models were put through eight test scenarios, all differing only on the image augmentation techniques used on the training data (and consequently the validation set. \Cref{tab:experiments} presents the results of these experiments.

    \begin{table}[ht]
        \centering
        \caption{Summary of the training experiments. T and V columns represent the minimum average loss on the training and validation sets of each model, respectively.}
        \label{tab:experiments}
        \begin{tabular}{|l|l|l|l|l|l|l|}
            \hline
            \textbf{Test} & \textbf{I3 T} & \textbf{I3 V} & \textbf{VGG T} & \textbf{VGG V} & \textbf{Eva T} & \textbf{Eva V} \\ \hline
            \#1           & 8.023         & 12.37         & 13.60          & 26.94          & 12.24          & 7.318          \\ \hline
            \#2           & 7.790         & 18.49         & 13.90          & 21.60          & 13.61          & 11.05          \\ \hline
            \#3           & 8.236         & 19.57         & 13.90          & 18.70          & 15.25          & 10.54          \\ \hline
            \#4           & 6.983         & 14.36         & 11.43          & 25.69          & 11.97          & 13.67          \\ \hline
            \#5           & 6.609         & 21.37         & 13.69          & 47.42          & 10.50          & 17.15          \\ \hline
            \#6           & 8.314         & 15.24         & 15.23          & 21.94          & 15.08          & 12.36          \\ \hline
            \#7           & 8.384         & 22.10         & 14.07          & 28.12          & 14.98          & 17.16          \\ \hline
            \#8           & 8.320         & 22.64         & 13.97          & 32.22          & 15.53          & 19.89          \\ \hline
        \end{tabular}
    \end{table}

    \begin{itemize}
        \item \textbf{Test \#1} No augmentation
        \item \textbf{Test \#2} Horizontal flips
        \item \textbf{Test \#3} Translations
        \item \textbf{Test \#4} Background images
        \item \textbf{Test \#5} Rotations
        \item \textbf{Test \#6} Flips and translations
        \item \textbf{Test \#7} Flips, translations and rotations
        \item \textbf{Test \#8} Flips, translations, rotations and backgrounds
    \end{itemize}

\begin{figure}[ht]
        \centering
        \includegraphics[width=0.48\textwidth]{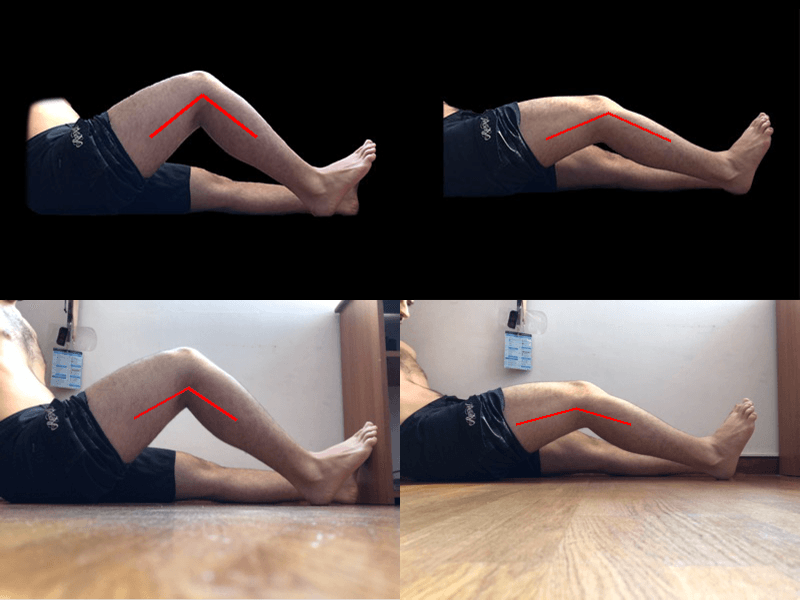}
        \caption{InceptionV3 predictions on real photographs. The red lines connect the three predicted points.}
        \label{fig:final-prediction}
    \end{figure}

    The results suggest a slight performance reduction on each test, as the number of transformations increases. This was expected, since more transformations result in a more diverse dataset, which hinders the task of detecting meaningful patterns across all samples. We can also see that VGG16 is the model with worst performance, while InceptionV3 and Eva produce similarly good results. In fact, even testing with unedited real pictures produces significantly good predictions, as seen in \cref{fig:final-prediction}.

\section{Conclusion} \label{sec:conclusion}
    This research proposes a novel solution for the problem of being able to accurately, easily and quickly measure the range of motion of an recovering patient's knee. The final step of implementing the solution in the form of a smartphone application is yet to be achieved. However, regarding the core challenge of the solution, image recognition through deep learning supported by synthetic data generation, the outcome was satisfactory. Results support the hypothesis that artificial data can effectively be used for this type of task, showing that the trained CNN models are capable of generalizing from previous experience with this kind of data to real-life examples. While current results are promising, further improvement is required before implementing the final solution. Future work might focus on enhancing the automated image recognition task by exploring a range of different strategies.
    For instance, instead of training a CNN to predict three key points, it could be worthy to predict just the coordinates of the knee along with two vectors representing the directions of the thigh and lower leg, or even the flexion angle directly. Additional CNN architectures should also be explored, as there is a great number of options available. Regarding the dataset, improving its diversity would be helpful, possibly by using several new 3D characters with different anatomies, clothes, etc. Once a final solution, including mobile application, is developed, it shall be tested on a clinical environment by health professionals on actual patients. Since this research provides insight on general image recognition techniques, it is not exclusive to the knee and might possibly be applied to additional types of measurements in the future.
%



\begin{thebibliography}{21}
\providecommand{\natexlab}[1]{#1}
\providecommand{\url}[1]{#1}
\csname url@samestyle\endcsname
\providecommand{\newblock}{\relax}
\providecommand{\bibinfo}[2]{#2}
\providecommand{\BIBentrySTDinterwordspacing}{\spaceskip=0pt\relax}
\providecommand{\BIBentryALTinterwordstretchfactor}{4}
\providecommand{\BIBentryALTinterwordspacing}{\spaceskip=\fontdimen2\font plus
\BIBentryALTinterwordstretchfactor\fontdimen3\font minus
    \fontdimen4\font\relax}
\providecommand{\BIBforeignlanguage}[2]{{%
\expandafter\ifx\csname l@#1\endcsname\relax
\typeout{** WARNING: IEEEtranN.bst: No hyphenation pattern has been}%
\typeout{** loaded for the language `#1'. Using the pattern for}%
\typeout{** the default language instead.}%
\else
\language=\csname l@#1\endcsname
\fi
#2}}
\providecommand{\BIBdecl}{\relax}
\BIBdecl

\bibitem[Majewski et~al.(2006)Majewski, Susanne, and Klaus]{Majewski:2006df}
M.~Majewski, H.~Susanne, and S.~Klaus, ``{Epidemiology of athletic knee
    injuries: A 10-year study},'' \emph{The Knee}, vol.~13, no.~3, pp. 184--188,
    Jun. 2006.

\bibitem[Edwards et~al.(2004)Edwards, Greene, Davis, Kovacik, Noe, and
    Askew]{Edwards:2004fg}
J.~Z. Edwards, K.~A. Greene, R.~S. Davis, M.~W. Kovacik, D.~A. Noe, and M.~J.
    Askew, ``{Measuring flexion in knee arthroplasty patients},'' \emph{The
    Journal of Arthroplasty}, vol.~19, no.~3, pp. 369--372, Apr. 2004.

\bibitem[Enwemeka(1986)]{Enwemeka:1986vy}
C.~S. Enwemeka, ``{Radiographic verification of knee goniometry.}''
    \emph{Scandinavian journal of rehabilitation medicine}, vol.~18, no.~2, pp.
    47--49, 1986.

\bibitem[Hellebrandt et~al.(1949)Hellebrandt, Duvall, and
    Moore]{Hellebrandt:1949je}
F.~A. Hellebrandt, E.~N. Duvall, and M.~L. Moore, ``{The Measurement of Joint
    Motion: Part III{\textemdash}Reliability of Goniometry*},'' \emph{Physical
    Therapy}, vol.~29, no.~7, pp. 302--307, Jul. 1949.

\bibitem[Roach et~al.(2013)Roach, San~Juan, Suprak, and Lyda]{Roach:2013wr}
S.~Roach, J.~G. San~Juan, D.~N. Suprak, and M.~Lyda, ``{Concurrent validity of
    digital inclinometer and universal goniometer in assessing passive hip
    mobility in healthy subjects.}'' \emph{International journal of sports
    physical therapy}, vol.~8, no.~5, pp. 680--688, Oct. 2013.

\bibitem[Mosa et~al.(2012)Mosa, Yoo, and Sheets]{Mosa:2012fk}
A.~S.~M. Mosa, I.~Yoo, and L.~Sheets, ``{A systematic review of healthcare
    applications for smartphones.}'' \emph{BMC medical informatics and decision
    making}, vol.~12, no.~1, p.~67, Jul. 2012.

\bibitem[Jenny(2013)]{Jenny:2013bw}
J.-Y. Jenny, ``{Measurement of the Knee Flexion Angle With a
    Smartphone-Application is Precise and Accurate},'' \emph{The Journal of
    Arthroplasty}, vol.~28, no.~5, pp. 784--787, May 2013.

\bibitem[Vercelli et~al.(2017)Vercelli, Sartorio, Bravini, and
    Ferriero]{Vercelli:2017dk}
S.~Vercelli, F.~Sartorio, E.~Bravini, and G.~Ferriero, ``{DrGoniometer: a
    reliable smartphone app for joint angle measurement},'' \emph{British Journal
    of Sports Medicine}, vol.~51, no.~23, pp. 1703--1704, Nov. 2017.

\bibitem[Ferriero et~al.(2013)Ferriero, Vercelli, Sartorio, Mu{\~n}oz~Lasa,
    Ilieva, Brigatti, Ruella, and Foti]{Ferriero:2013iy}
G.~Ferriero, S.~Vercelli, F.~Sartorio, S.~Mu{\~n}oz~Lasa, E.~Ilieva,
    E.~Brigatti, C.~Ruella, and C.~Foti, ``{Reliability of a smartphone-based
    goniometer for knee joint goniometry.}'' \emph{International journal of
    rehabilitation research. Internationale Zeitschrift fur
    Rehabilitationsforschung. Revue internationale de recherches de
    readaptation}, vol.~36, no.~2, pp. 146--151, Jun. 2013.

\bibitem[Otter et~al.(2015)Otter, Agalliu, Baer, Hales, Harvey, James, Keating,
    McConnell, Nelson, Qureshi, Ryan, St~John, Waddington, Warren, and
    Wong]{Otter:2015gm}
S.~J. Otter, B.~Agalliu, N.~Baer, G.~Hales, K.~Harvey, K.~James, R.~Keating,
    W.~McConnell, R.~Nelson, S.~Qureshi, S.~Ryan, A.~St~John, H.~Waddington,
    K.~Warren, and D.~Wong, ``{The reliability of a smartphone goniometer
    application compared with a traditional goniometer for measuring first
    metatarsophalangeal joint dorsiflexion.}'' \emph{Journal of foot and ankle
    research}, vol.~8, no.~1, p.~30, 2015.

\bibitem[Langley(1988)]{Langley:1988jk}
P.~Langley, ``{Machine learning as an experimental science},'' \emph{Machine
    Learning}, vol.~3, no.~1, pp. 5--8, 1988.

\bibitem[Sinha et~al.(2018)Sinha, Pandey, and Pattnaik]{Sinha:2018vm}
R.~K. Sinha, R.~Pandey, and R.~Pattnaik, ``{Deep Learning For Computer Vision
    Tasks: A review},'' Apr. 2018.

\bibitem[Voulodimos et~al.(2018)Voulodimos, Doulamis, Doulamis, and
    Protopapadakis]{Voulodimos:2018bf}
A.~Voulodimos, N.~Doulamis, A.~Doulamis, and E.~Protopapadakis, ``{Deep
    Learning for Computer Vision: A Brief Review.}'' \emph{Computational
    intelligence and neuroscience}, vol. 2018, p. 7068349, 2018.

\bibitem[Cao et~al.(2018)Cao, Wang, Ming, and Gao]{Cao:2018gb}
W.~Cao, X.~Wang, Z.~Ming, and J.~Gao, ``{A review on neural networks with
    random weights},'' \emph{Neurocomputing}, vol. 275, pp. 278--287, Jan. 2018.

\bibitem[Pan and Yang(2010)]{Pan:2010dm}
S.~J. Pan and Q.~Yang, ``{A Survey on Transfer Learning},'' \emph{IEEE
    Transactions on Knowledge and Data Engineering}, vol.~22, no.~10, pp.
    1345--1359, Aug. 2010.

\bibitem[Shin et~al.(2016)Shin, Roth, Gao, Lu, Xu, Nogues, Yao, Mollura, and
    Summers]{Shin:2016cx}
H.-C. Shin, H.~R. Roth, M.~Gao, L.~Lu, Z.~Xu, I.~Nogues, J.~Yao, D.~Mollura,
    and R.~M. Summers, ``{Deep Convolutional Neural Networks for Computer-Aided
    Detection: CNN Architectures, Dataset Characteristics and Transfer
    Learning},'' \emph{IEEE Transactions on Medical Imaging}, vol.~35, no.~5, pp.
    1285--1298, Feb. 2016.

\bibitem[Keivan~Ekbatani et~al.()Keivan~Ekbatani, Pujol, and
    Segui]{KeivanEkbatani:gn}
H.~Keivan~Ekbatani, O.~Pujol, and S.~Segui, ``{Synthetic Data Generation for
    Deep Learning in Counting Pedestrians},'' in \emph{6th International
    Conference on Pattern Recognition Applications and Methods}.\hskip 1em plus
    0.5em minus 0.4em\relax SCITEPRESS - Science and Technology Publications, pp.
    318--323.

\bibitem[Jegou et~al.(2008)Jegou, Douze, and Schmid]{Jegou:2008il}
H.~Jegou, M.~Douze, and C.~Schmid, ``{Hamming Embedding and Weak Geometric
    Consistency for Large Scale Image Search},'' in \emph{Computer Vision
    {\textendash} ECCV 2008}.\hskip 1em plus 0.5em minus 0.4em\relax Berlin,
    Heidelberg: Springer, Berlin, Heidelberg, Oct. 2008, pp. 304--317.

\bibitem[Szegedy et~al.()Szegedy, Vanhoucke, Ioffe, Shlens, and
    Wojna]{Szegedy:cv}
C.~Szegedy, V.~Vanhoucke, S.~Ioffe, J.~Shlens, and Z.~Wojna, ``{Rethinking the
    Inception Architecture for Computer Vision},'' in \emph{2016 IEEE Conference
    on Computer Vision and Pattern Recognition (CVPR)}.\hskip 1em plus 0.5em
    minus 0.4em\relax IEEE, pp. 2818--2826.

\bibitem[Simonyan and Zisserman(2014)]{Simonyan:2014ws}
K.~Simonyan and A.~Zisserman, ``{Very Deep Convolutional Networks for
    Large-Scale Image Recognition.}'' \emph{CoRR}, vol. abs/1409.1556, 2014.

\bibitem[Wichrowska et~al.(2017)Wichrowska, Maheswaranathan, Hoffman,
    Colmenarejo, Denil, de~Freitas, and Sohl-Dickstein]{Wichrowska:2017vo}
O.~Wichrowska, N.~Maheswaranathan, M.~W. Hoffman, S.~G. Colmenarejo, M.~Denil,
    N.~de~Freitas, and J.~Sohl-Dickstein, ``{Learned Optimizers that Scale and
    Generalize},'' \emph{arXiv.org}, Mar. 2017.

\end{thebibliography}
\end{document}